\documentclass[lettersize,journal]{IEEEtran}
\usepackage{amsmath,amsfonts}
\usepackage{array}
\usepackage{textcomp}
\usepackage{stfloats}
\usepackage{url}
\usepackage{verbatim}
\usepackage{cite}
\ifCLASSINFOpdf
  \usepackage[pdftex]{graphicx}
  \DeclareGraphicsExtensions{.eps,.pdf,.jpeg,.png}
\else
  \usepackage[dvips]{graphicx}
  \DeclareGraphicsExtensions{.eps,.pdf}
\fi
\ifCLASSOPTIONcompsoc
  \usepackage[caption=false,font=normalsize,labelfont=sf,textfont=sf]{subfig}
\else
  \usepackage[caption=false,font=footnotesize]{subfig}
\fi

\usepackage{amssymb}
\usepackage{bm}
\usepackage{xfrac}
\usepackage{mathrsfs}

\graphicspath{{Figures/}{second_figs}}
\usepackage{ulem}

\usepackage{booktabs}
\usepackage{threeparttable}
\usepackage{multirow}

\usepackage[ruled,linesnumbered]{algorithm2e}

\DeclareMathOperator*{\argmax}{argmax}

\def\opl{\mathfrak{l}}
\def\opt{\mathfrak{t}}

\def\opA{\mathfrak{A}}

\def\caLo{\mathcal{L}_\text{original}}
\def\caLt{\mathcal{L}_\text{target}}
\def\naDo{\nabla_\text{original}}
\def\naDt{\nabla_\text{target}}
\def\noise{\bm{\delta}}
\def\patch{\mathbf{p}}
\def\eg{e.g.,}
\def\ie{i.e.,}
\def\aka{aka.}
\def\etal{\textit{et al.}}
\def\entitle{Adversarial Patch Attacks and Defences in Vision-Based Tasks: A Survey}
\def\journal{IEEE Transactions on}% Cybernetics}
\let\emph\textit
\newcommand{\tabincell}[2]{\begin{tabular}{@{}#1@{}}#2\end{tabular}}
\newcolumntype{C}[1]{>{\centering\arraybackslash}p{#1}}
\newcolumntype{L}[1]{>{\raggedright\arraybackslash}p{#1}}
\newcolumntype{R}[1]{>{\raggedleft\arraybackslash}p{#1}}

\usepackage{tikz}
\usetikzlibrary{calc,matrix,decorations.markings,decorations.pathreplacing}
\tikzset{ 
  table/.style={
    matrix of nodes, nodes in empty cells,
    row sep=-\pgflinewidth, column sep=-\pgflinewidth,
    nodes={rectangle,text width=2.1cm,align=center},
    text depth=.8ex, text height=1.7ex, 
  }}
\newcommand{\cbox}[1]{\parbox[t]{2cm}{\centering #1}}
\definecolor{clrB}{HTML}{F0B775}
\definecolor{clrE}{HTML}{2E94B9}

\begin{document}

% \title{A Sample Article Using IEEEtran.cls\\ for IEEE Journals and Transactions}
\title{\entitle{}}

\author{%
% Abhijith Sharma,$^*$ Yijun Bian,$^*$ Phil Munz, and Apurva Narayan,~\emph{Senior Member,~IEEE}
Abhijith Sharma$^*$, Yijun Bian$^*$, Phil Munz, and Apurva Narayan,~\emph{Senior Member,~IEEE}
\thanks{Manuscript received June 12, 2022. A. Sharma and Y. Bian contributed equally to this work. % \textcolor{red}{TODO}
% \emph{(Corresponding author: Apurva Narayan.)}
\emph{(Corresponding author: Yijun Bian.)}
}\thanks{
A. Sharma and A. Narayan are with the Department of Computer Science, University of British Columbia, Kelowna, BC V1V 1V7, Canada (E-mails: sharma86@mail.ubc.ca; apurva.narayan@ubc.ca). %
% }\thanks{% School of Engineering % Abhijith Sharma and Apurva Narayan
Y. Bian is with the School of Computer Science and Technology, University of Science and Technology of China, Hefei, Anhui 230027, China (E-mail: yjbian@mail.ustc.edu.cn). %
% }\thanks{
P. Munz is with TrojAI Inc., Saint John, NB E2L 1G2, Canada (E-mail: phil.munz@troj.ai).}% New Brunswick % the
\thanks{This work has been submitted to the IEEE for possible publication. Copyright may be transferred without notice, after which this version may no longer be accessible.}
}
\iffalse%
\author{IEEE Publication Technology,~\IEEEmembership{Staff,~IEEE,}
        % <-this % stops a space
\thanks{This paper was produced by the IEEE Publication Technology Group. They are in Piscataway, NJ.}% <-this % stops a space
\thanks{Manuscript received April 19, 2021; revised August 16, 2021.}}
\fi%

\markboth{\journal{},~Vol.~, No.~, Month~2022}%
{Author \MakeLowercase{\textit{et al.}}: \entitle{}}


\maketitle

\begin{abstract}
Adversarial attacks in deep learning models, especially for safety-critical systems, are gaining more and more attention in recent years, due to the lack of trust in the security and robustness of AI models. 
Yet the more primitive adversarial attacks might be physically infeasible or require some resources that are hard to access like the training data, which motivated the emergence of patch attacks. 
In this survey, we provide a comprehensive overview to cover existing techniques of adversarial patch attacks, aiming to help interested researchers quickly catch up with the progress in this field. 
We also discuss existing techniques for developing detection and defences against adversarial patches, aiming to help the community better understand this field and its applications in the real world. 
% in security concerns. 
\end{abstract}

\begin{IEEEkeywords}
Deep learning, adversarial attack, patch attack, patch detection, patch defence.
% Article submission, IEEE, IEEEtran, journal, \LaTeX, paper, template, typesetting.
\end{IEEEkeywords}

% match that (performance) of human % match (the) human performance
% Apart from a general safety threat,  the attacks could also help in 
\section{Introduction}
\IEEEPARstart{D}{eep} learning techniques have been rapidly developed with a tremendous increase in not only performance but also applicability over the last two decades. 
Specifically, the advent of convolution-based neural networks (CNNs) led to a giant leap of progress on vision-based tasks such as face recognition and object recognition \cite{lawrence1997face,nebauer1998evaluation,lecun1999object}. 
These various CNNs of multifarious architectures have gained huge popularity and extraordinary performance that could match the human performance (like perception and decision making) and even more \cite{krizhevsky2012imagenet,khan2020survey,he2015delving,sabour2017dynamic,tian2017towards}. 
However, the adoption of CNN-based systems for real-world applications has not been closely consistent with their encouraging performance improvement, with multi-fold reasons---including the lack of explainability \cite{rudin2019stop,dam2018explainable,angelov2020towards}, the infeasibility of implementation \cite{ravi2016deep,botev2017practical}, limited training and slow response time \cite{chen2021refit,oh2020deep,anthony2017thinking}---but most essentially, due to the lack of trust in security concerns \cite{karmakar2021assessing,li2020trustworthy,kok2020trust,hassan2020robust,von2021transparency}, especially for safety-critical systems. 
Quite recently, a myriad of adversarial attacks demonstrated the vulnerability of deep learning systems, exposing the threat of implementing them in the real world \cite{kurakin2018adversarial,moosavi2016deepfool,carlini2017towards,papernot2017practical,athalye2018synthesizing,madry2018towards,su2019one}. 
Apart from a general security threat, the attacks also help in understanding the limitations of CNN-based models in some cases \cite{papernot2016limitations,karmon2018lavan}.

The most primitive adversarial attacks typically demonstrated in CNNs misled the model imperceptibly via small additive noise \cite{szegedy2014intriguing,goodfellow2014explaining}. 
Some attacks that followed often overlapped the main object in the scene, attacking the model in a natural way \cite{zhu2019robustness,kortylewski2020combining}. 
Meanwhile, a new kind of attack termed Poisoning attacks misled the models by feeding them incorrect patterns during the training, such as Poison Frogs \cite{shafahi2018poison} and Backdoor attacks \cite{chen2017targeted}. 
Despite their efficiency, most of these attacks were infeasible either physically or due to the required resources, for example, access limitations to the training data. 
Therefore, a form of localised and visible contiguous perturbation of image pixels emerged, known as \emph{Patch attacks} \cite{brown2017adversarial}. 
A patch is a patterned sub-image that is generally masked over the input image to attack the model, and its inherent characteristic makes it physically implementable in a real-world scenario. 
Moreover, patch attacks are universal and evasive, which allows an adversary to run even when it has no knowledge about the system or the used training data \cite{co2021real,zolfi2021translucent,tran2012evasive,edwards2015censoring}. 
The only limitation of patch attacks is the visibility to human eyes, although humans are seldom involved in most AI-based systems.

The detection and defence of practical patch attacks have gained the utmost popularity in the community concerning the security threat of CNNs. 
However, most of the existing surveys around adversarial attacks \cite{akhtar2018threat,akhtar2021advances,xu2020adversarial,chakraborty2018adversarial,bhambri2019survey,sun2018survey,zhang2021survey,ren2020adversarial,serban2020adversarial,machado2021adversarial} lack a dedicated focus on adversarial patch attacks. 
To bridge this gap, we comprehensively summarise various adversarial patches and the corresponding mitigation techniques in this survey, presenting a start for the research in similar lines of practical attacks for vision-based tasks. 
The remainder of this article is organised as follows. 
To begin with, we briefly introduce the necessary definitions and notations in Section~\ref{sec: definition_of_terms}. 
After that, we specifically report summaries of adversarial patch attacks in Section~\ref{sec: patch_attacks} and that of patch detection/defence in Section~\ref{sec: patch_defense}, respectively. 
In the end, we discuss some opening challenges of patch attacks for future work in Section~\ref{sec: conclusion}.

\section{Preliminaries}
\label{sec: definition_of_terms}

In this section, we formally define the necessary notations and introduce relative concepts that are used in this paper. 

% \paragraph*{Notations:} 
\textbf{Notations}\indent%
In this paper, tensors, vectors, and scalars are denoted by bold italic lowercase letters (\eg{} $\bm{x}$), bold lowercase letters (\eg{} $\mathbf{x}$), and italic lowercase letters (\eg{} $x$), respectively. 
Then the transpose of a vector is denoted by $\mathbf{x}^\mathsf{T}$. 
Data/hypothesis spaces and distributions are represented by bold script uppercase letters (\eg{} $\mathcal{X}$) and serif uppercase letters (\eg{} $\mathsf{X}$), respectively. 
Operators are denoted by particular alphabet symbols (\eg{} $\opA$). 
% Finally, 
We use $\mathbb{R}$ and $\mathbb{E}$ to denote the real space and the expectation of a random variable, respectively. 
Other used symbols and their definitions are summarised in Table~\ref{tab:notations}. 
Then we follow these notations to formulate the problem and introduce necessary concepts in the meanwhile.

% \iffalse%
\begin{table}[ht]
\centering
\caption{Used notations and the corresponding definitions in this paper}
\label{tab:notations}
\scalebox{1.}{
\begin{tabular}{lp{.374\textwidth}}%
  \toprule
  Notation & Definition \\
  \midrule %
  $\bm{x}\in\mathcal{X}$ & the input of neural networks where $\mathcal{X}=\mathbb{R}^{n_d} =\mathbb{R}^{w\cdot h\cdot c}$ with $w,h,$ and $c$ as the width, height, and the number of channels ($c=3$ in RGB images) \\
  $y\in\mathcal{Y}$ & the output label produced by a generic model $\mathcal{M}$ for the input image, where $\mathcal{Y}= \{0,1,...,n_c-1\}$ and $n_c$ denotes the number of labels \\
  $\vec{y}$ & the membership probability vector, predicted by the pre-trained model $\mathcal{M}(y|\bm{x})$ for a given image input $\bm{x}$ \\
  $\bm{x}' \in\mathcal{X}$ & the perturbed image given the crafted noise $\noise\in [0,1]^{whc}$ \\
  $\patch\in\mathcal{P}$ & the binary pixel block to mask a small restricted region, where $\mathcal{P}\subset\{0,1\}^{wh}$ \\
  \bottomrule
\end{tabular}
}
\end{table}
% \fi%

There are lots of attacks in the literature and many ways to classify them as well. 
For example, the demand for the information of the target network distinguishes between the \textit{white-box} and \textit{black-box attack}. 
In the former, the attack has full access to the network information (including the architecture, parameters, gradients, etc), while none of the model's inner configuration would be available in the latter. 
Sometimes, one can create an attack called a \textit{gray-box attack} which trains a generative model to gain the ability to produce adversarial examples without a victim model, in which case the victim model would be needed before the generative model was trained good enough. 
A crafted attack would be called an \textit{evasive attack} if it is carried out to fool a model at test time without any previous knowledge of the model's architecture or parameters. 
Note that a \textit{universal attack} means its performance would be independent of the image or the model, knowing from the image-specific or model-specific attack. 
% A \emph{universal attack} means that

% \begin{figure}[ht]
\begin{figure*}[th]
\centering
\includegraphics[width=.9\textwidth]{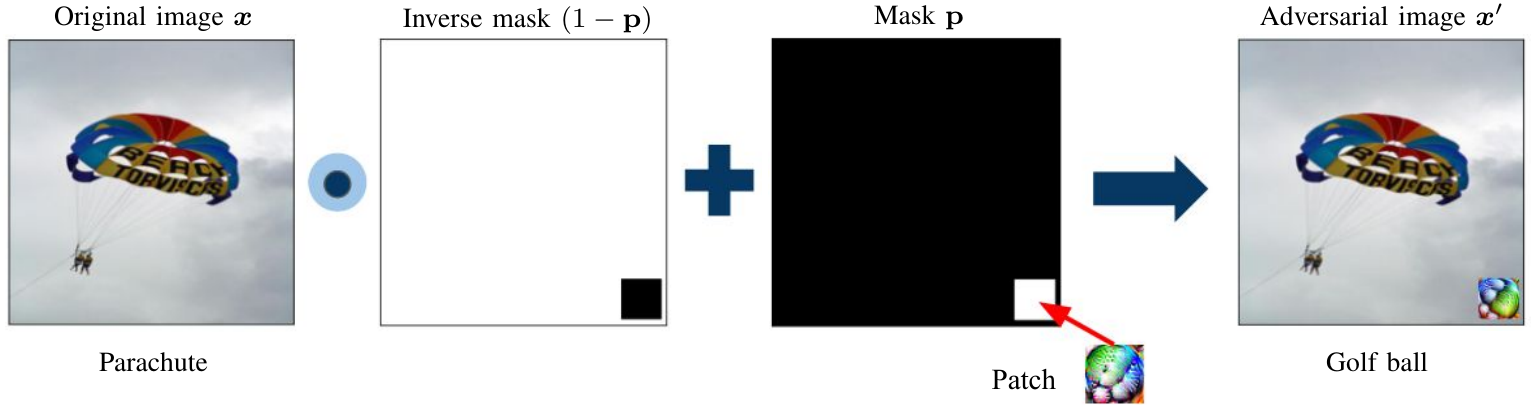}
\hspace{3mm}\vspace{-2mm}
\caption{% \centering 
Adversarial patch attack procedure (white is $1$ and black is $0$). 
The adversarial example is generated by $\bm{x}'= (1-\mathbf{p})\odot\bm{x} +\mathbf{p}\odot\mathbf{\delta}$, where $\mathbf{\delta}$ and $\mathbf{p}$ are the adversarial patch noise and the adversarial patch, respectively. Note that $\odot$ represents the Hadamard operator performing element-wise multiplication. 
}\label{fig:attack_proc}
\end{figure*}
% \end{figure}

Let $\bm{x}\in\mathcal{X}$ be an original RGB image as input, referred to as the \textit{clean image}, and $\bm{x}'= \bm{x}+\bm{\delta}$ be the perturbed image for the crafted noise $\bm{\delta}$. 
Attacks requiring access to every pixel in the image would be called a \textit{digital attack} and is usually infeasible in real-world scenarios. 
For a generic model $\mathcal{M}$, the membership probability vector $\vec{y}$ for a given image $\bm{x}$ is predicted by its pre-trained version $\mathcal{M}(y|\bm{x})$. 
Then the original label $y_o$ and target label $y_t$ could be formulated as
\begin{small}
\begin{subequations}
\begin{align}
  y_o&= \argmax[ \mathcal{M}(\vec{y}|\bm{x}) ] \,,\\
  y_t&= \argmax[ \mathcal{M}(\vec{y}|\bm{x}') ] \,,
\end{align}%
\label{eq:1}%
\end{subequations}%
\end{small}%
predicted on the original image $\bm{x}$ and perturbed image $\bm{x}'$, respectively. 
Attacks deceiving the model to predict all images to one predefined label (\ie{} a target label) would be called a \textit{targeted attack}, while those misleading the model to any labels other than the true label of this image would be called an \textit{untargeted attack}. 

Additionally, attacks that force the model to learn inaccurate patterns based on the manipulated training data would be called a \textit{poisoning attack}. 
Another point is that the visibility to a human's eye distinguishes between the \textit{imperceptible} and \textit{perceptible/visible attack} when the perturbations are added to an image to cheat the model. 
Note that the metric \textit{transferability} is used to measure the generalisation of an attack, describing its capability to fool a model on an image from one dataset where neither the model nor the dataset may be a necessary part of its training.

Crafting an attack typically means formulating an optimisation problem and utilising a gradient descent based algorithm to iteratively train the additive noise on the image. 
In an untargeted attack, the probability of classifying $\bm{x}$ as the original label $\mathcal{M}(y=y_o|\bm{x}')$ is minimised, while in a targeted attack, the probability of classifying $\bm{x}$ as the target label $\mathcal{M}(y=y_t|\bm{x}')$ is maximised. 
In norm-based imperceptible adversarial attacks, the noise $\noise$ is added to each pixel over the entire image, given by $\bm{x}'=\bm{x}+\noise$. 
In contrast, the perturbed region is confined to a small restricted region in patch attacks, defined as
\begin{equation}
  \small
  \bm{x}' = 
  (1-\patch)\odot \bm{x}
  +\patch\odot\noise 
  \,,\label{eq:2}%
\end{equation}%
where $\noise$ is the adversarial patch noise and $\patch$ represents the binary pixel block to mask the patch area, known as an \textit{adversarial patch}. 
Note that the symbol $\odot$ represents the Hadamard operator that performs element-wise multiplication of pixels from the respective matrices. 
The pixel block (mask) determines the area and location of the patch over the image. 
Figure~\ref{fig:attack_proc} shows an example of a perturbed image (\ie{} the \textit{adversarial image}), where the attacker that crafts the adversarial patch is known as the \textit{adversary}.

\iffalse%
\begin{algorithm}[H]
\caption{Weighted Tanimoto ELM.}\label{alg:alg1}
\begin{algorithmic}
\STATE 
\STATE {\textsc{TRAIN}}$(\mathbf{X} \mathbf{T})$
\STATE \hspace{0.5cm}$ \textbf{select randomly } W \subset \mathbf{X}  $
\STATE \hspace{0.5cm}$ N_\mathbf{t} \gets | \{ i : \mathbf{t}_i = \mathbf{t} \} | $ \textbf{ for } $ \mathbf{t}= -1,+1 $
\STATE \hspace{0.5cm}$ B_i \gets \sqrt{ \textsc{max}(N_{-1},N_{+1}) / N_{\mathbf{t}_i} } $ \textbf{ for } $ i = 1,...,N $
\STATE \hspace{0.5cm}$ \hat{\mathbf{H}} \gets  B \cdot (\mathbf{X}^T\textbf{W})/( \mathbb{1}\mathbf{X} + \mathbb{1}\textbf{W} - \mathbf{X}^T\textbf{W} ) $
\STATE \hspace{0.5cm}$ \beta \gets \left ( I/C + \hat{\mathbf{H}}^T\hat{\mathbf{H}} \right )^{-1}(\hat{\mathbf{H}}^T B\cdot \mathbf{T})  $
\STATE \hspace{0.5cm}\textbf{return}  $\textbf{W},  \beta $
\STATE 
\STATE {\textsc{PREDICT}}$(\mathbf{X} )$
\STATE \hspace{0.5cm}$ \mathbf{H} \gets  (\mathbf{X}^T\textbf{W} )/( \mathbb{1}\mathbf{X}  + \mathbb{1}\textbf{W}- \mathbf{X}^T\textbf{W}  ) $
\STATE \hspace{0.5cm}\textbf{return}  $\textsc{sign}( \mathbf{H} \beta )$
\end{algorithmic}
\label{alg1}
\end{algorithm}
\fi%

%\iffalse%
\begin{algorithm*}
\small
\caption{Localised Noising Procedure}
\label{alg:seq}%
\KwIn{Image $\bm{x}$, model $\mathcal{M}$, target label $y_t$, target probability $s$, maximal number of iterations $n_t$}%
\BlankLine%
% \Begin{
$y_o= \argmax[\mathcal{M}(\vec{y}|\bm{x})]$ \;
$\noise= \vec{0}$ \;
$\bm{x}'= \opA\big( (1-\patch)\odot\bm{x}+\patch\odot\noise \big)$ 
\tcc*{\footnotesize EOT with patch operator $\opA$}
\While{$\mathcal{M}(y=y_t|\bm{x}') \geqslant s$ \&\& $i\leqslant n_t$}{
  $\vec{y}= \mathcal{M}(\bm{x})$ \;
  $\caLt= \ell(\vec{y},y_t)= \mathcal{M}(y=y_t|\bm{x}')$ %\tcp*{\small Loss function for the targeted attack}
  \tcc*{\footnotesize Loss function for the targeted attack}
  $\caLo= \ell(\vec{y},y_o)= \mathcal{M}(y=y_o|\bm{x}')$ %\tcp*{\small Loss function for untargeted attack}
  \tcc*{\footnotesize Loss function for untargeted attack}
  $\naDt= \frac{\partial}{\partial \bm{x}} \caLt$ %\tcp*{\small Calculating gradient w.r.t image pixels}
  \tcc*{\footnotesize Calculating gradient w.r.t image pixels}
  $\naDo= \frac{\partial}{\partial \bm{x}} \caLo$ \;
  $\noise= \noise- \epsilon\cdot  \naDt$ %\tcp*{\small Update pixels to minimise loss in the targeted attack}
  \tcc*{\footnotesize Update pixels to minimise loss in the targeted attack}
  $\noise= \noise+ \epsilon\cdot \naDo$ %\tcp*{\small Update pixels to maximise loss in the untargeted attack}
  \tcc*{\footnotesize Update pixels to maximise loss in the untargeted attack}
  $i= i+1$ \;
  $\bm{x}'= (1-\patch)\odot\bm{x} +\patch\odot\noise$ %\tcp*{\small Produce the adversarial image}
  \tcc*{\footnotesize Produce the adversarial image}
}
% }% Begin
\end{algorithm*}
%\fi%

The region of pixel manipulation distinguishes between the \textit{global} and \textit{local perturbations}. 
In the former, every pixel in the image is manipulated to perform an attack, while in the latter, only pixels within a small restricted area are manipulated. 
If the perturbed pixels are all adjacent to each other, it would be a \textit{continuous perturbation} where one pixel can join all perturbed pixels without skipping one in between. 
By physically printing patches in the form of stickers or posters, the attack could also be applied to the real world, known as a \textit{physical attack}. 
Moreover, the mechanism to detect a patch's presence and location in an image is known as \textit{patch detection}, consequently, the mechanism to mitigate or nullify the effect of an adversarial patch attack on predictions is known as a \textit{patch defence}.

While different patch variants may vary greatly, they share a typical patch training methodology to form the basis, as shown in Algorithm~\ref{alg:seq}. 
In each iteration, the patch is applied to random locations in the image, ensuring that the trained patches are robust when applied to different locations. 
% Besides, an Expectation over Transformer (EOT) 
An Expectation over Transformation (EOT) \cite{athalye2018synthesizing} technique is often adopted as well to improve the robustness of the crafted patch, defined as 
\begin{equation}
  \small
  \hat{\patch}_{\opt} =
  \argmax_{\patch}
  \mathbb{E}_{
    \bm{x}\sim\mathsf{X},
    \opl\sim\mathsf{L}, 
    \opt\sim\mathsf{T}
  } [
    \log \mathcal{M}(
      \hat{y}| \opA(
        \patch, %, 
        \bm{x}, \opl, \opt
      ))] \,,\label{eq:3}%
\end{equation}%
where $\opA(\patch,\bm{x},\opl,\opt)$ is a patch application operator, applying the transformation $\opt$ (\eg{} scale or rotations) to the patch $\patch$ and then puts it at the location $\opl$ on the image $\bm{x}$ \cite{brown2017adversarial}. 
Note that $\mathsf{X}$ represents the distribution of training images, while $\mathsf{T}$ and $\mathsf{L}$ denote the distributions of random transformations and locations of the patch over the image, respectively. 
EOT allows the robustness of the patch to be extended to transformations like rotation and scaling. 
It is worth mentioning that various stopping criteria vary in the threshold depending on the quality of the crafted patch or the speed of training, such as the desired confidence score on the target label, the maximum number of iterations, or the timing when the objective function fails to decrease substantially (like meeting some threshold).

\section{Adversarial patch attacks}
\label{sec: patch_attacks}

Adversarial patch attacks are a class of localised perturbations (typically of contiguous pixels) that are capable of misleading the model in vision-based tasks. 
Unlike globally perturbed and imperceptible $\mathrm{L}_1, \mathrm{L}_2$, or $\mathrm{L}_\infty$ norm-based attacks, patch attacks are overt to human's eyes because of the inherent characteristic of being restricted to a small region. 
The local perturbations sacrifice stealth as they have to be large enough in magnitude to carry out a successful attack.

Despite the crucial covertness for an attacker, patch attacks belong to both the most straightforward physically attainable adversarial attacks and evasive attacks that require no access to the training data, therefore becoming particularly attractive in recent years \cite{geiping2021doesn,akhtar2021advances}. 
Patch attacks are usually printed in the form of posters or stickers, capable of being applied over the target objects in the scene. 
Robust patch stickers on the target object would not affect their attacking ability and effectiveness by locations. 
% by location
% realisable

In this section, we will summarise existing strategies to design patches for various scenarios and review the crafting mechanism for various patch attacks. 

\begin{figure*}[t]
\centering
% \caption{Highlights of the advances in adversarial attacks (The size of circle represents the relative number of papers published in that year)}
% \includegraphics[scale=.91]{/second_timeline.pdf}
% scale=.87
%
%
% \iffalse% \scalebox{1.}{%
% \begin{minipage}{.95\textwidth}
% \centering
\tikzset{global scale/.style={
  scale=#1,
  every node/.append style={scale=#1}
}}
\begin{tikzpicture}[global scale = .88]% .91
% \begin{tikzpicture}[scale=.74]% .87
% \tikzset{every node}=[font=\small,scale=0.74]
  \matrix (mat) [table] { % , text width=2.4cm
    |[fill=clrB!20]|      & |[fill=clrB!20]|  & |[fill=clrB!20]|
      & |[fill=clrB!20]|  & |[fill=clrB!20]|  & |[fill=clrB!20]|  &                   \\
    |[fill=clrB!30]|      & |[fill=clrB!30]|  & |[fill=clrB!30]|
      & |[fill=clrB!30]|  & |[fill=clrB!30]|  & |[fill=clrB!30]|  &                   \\
    |[fill=clrB!40]|       & |[fill=clrB!40]|   & |[fill=clrB!40]|
      & |[fill=clrB!40]|   & |[fill=clrB!40]|   & |[fill=clrB!40]| & |[fill=clrB!40]|   \\
    |[fill=clrB!50]|     & |[fill=clrB!50]| & |[fill=clrB!50]|
      & |[fill=clrB!50]| & |[fill=clrB!50]| & |[fill=clrB!50]|   &    |[fill=clrB!50]| \\
    % % % %
    |[fill=clrE!30]|       & |[fill=clrE!50]|   & |[fill=clrE!70]|
      & |[fill=clrE!80]|   & |[fill=clrE!90]|   & |[fill=clrE!90]| & |[fill=clrE!90]|   \\
    % |[fill=clrE!20]|       & |[fill=clrE!40]|   & |[fill=clrE!60]|
    %   & |[fill=clrE!70]|   & |[fill=clrE!80]|   & |[fill=clrE!80]| & |[fill=clrE!80]|   \\
    |[fill=clrE!30]|       & |[fill=clrE!50]|   & |[fill=clrE!70]|
      & |[fill=clrE!80]|   & |[fill=clrE!90]|   & |[fill=clrE!90]| & |[fill=clrE!90]|   \\
    |[fill=clrE!30]|       & |[fill=clrE!50]|   & |[fill=clrE!70]|
      & |[fill=clrE!80]|   & |[fill=clrE!90]|   & |[fill=clrE!90]| &                   \\
    |[fill=clrE!30]|       & |[fill=clrE!50]|   & |[fill=clrE!70]|
      & |[fill=clrE!80]|   & |[fill=clrE!90]|   & |[fill=clrE!90]| &                   \\
  };

  % horizontal rules
  \foreach \row in {2,3,4}
    \draw[white] (mat-\row-1.north west) -- (mat-\row-6.north east);
  \draw[white,ultra thick] (mat-1-1.north west) -- (mat-1-7.north east);
  \draw[white,ultra thick] (mat-5-1.north west) -- (mat-5-7.north east);

  % vertical rules
  \foreach \col in {2,3,4,5}
    \draw[white] (mat-5-\col.north west) -- (mat-8-\col.south west);

  % The labels
  \iffalse%
  \node[fill=colfour] at (mat-1-3) {\small First for classification tasks \cite{37} and object detection \cite{59}}; % [37], [59]
  \node[fill=colfive] at (mat-2-3) {\small First GAN-based \cite{55}, physical patch \cite{67}, and to fool cameras \cite{69--70}}; % [55], [67], [69-70]
  \node[fill=colsix] at (mat-3-3) {\small First for attacking depth estimation \cite{102} and image semantics \cite{89}}; % [102], [89]
  \node[fill=colseven] at (mat-4-3) {\small Data independent adversarial patch \cite{58}, and adaptive adversarial patch \cite{71}};
  % [58], [71]
  \fi%
  
  % xshift=3em /2em
  \node[xshift=1.6em, fill=clrB!20] at (mat-1-3) {\small First for classification tasks \cite{brown2017adversarial} and object detection \cite{liu2018dpatch}}; % [37], [59]
  \node[xshift=1.6em, fill=clrB!30] at (mat-2-3) {\small First GAN-based \cite{liu2019perceptual}, physical patch \cite{song2018physical}, and to fool cameras \cite{thys2019fooling,den2020adversarial}}; % [55], [67], [69--70]
  \node[xshift=1.6em, fill=clrB!40] at (mat-3-3) {\small First for attacking depth estimation \cite{yamanaka2020adversarial} and image semantics \cite{mirsky2021ipatch}}; % [102], [89]
  \node[xshift=1.6em, fill=clrB!50] at (mat-4-3) {\small Data independent adversarial patch \cite{zhou2021data}, and adaptive adversarial patch \cite{lu2021scale}}; % [58], [71]
  
  \iffalse%
  \node at ([yshift=-10pt]mat-6-1) {\cbox{\small \mbox{~}\vspace{2.mm} \\\mbox{[37]} \\\mbox{~} \\\mbox{~} \\\normalsize \vspace{2mm}\textbf{in 2017} }}; %
  \node at ([yshift=-10pt]mat-6-2) {\cbox{\small \mbox{~}\vspace{2mm} \\\mbox{DPatch [59],}\\\mbox{and [67]} \\\mbox{~} \\\normalsize \vspace{2mm}\textbf{in 2018} }}; %
  \node at ([yshift=-10pt]mat-6-3) {\cbox{\small \mbox{~}\vspace{2mm} \\\mbox{PS-GAN [55],}\\\mbox{and [69]} \\\mbox{~} \\\normalsize \vspace{2mm}\textbf{in 2019} }}; %
  \node at ([yshift=-10pt]mat-6-4) {\cbox{\small \mbox{~}\vspace{2.mm} \\\mbox{[70, 102]} \\\mbox{~} \\\mbox{~} \\\normalsize \vspace{2mm}\textbf{in 2020} }}; %
  \fi%
  
  \node at ([yshift=-10pt, xshift=13pt]mat-6-5) {\cbox{\small \mbox{DiAP \cite{zhou2021data},}\\\mbox{Patch-Noobj \cite{lu2021scale},}\\\mbox{RAP (IPatch) \cite{mirsky2021ipatch}} \\\normalsize \vspace{2mm}\textbf{in 2021} }}; % [58], [71], [89]
  % xshift=13pt
  % \node at ([yshift=-10pt]mat-6-5) {\cbox{\small \mbox{DiAP [58], ~}\\\mbox{Patch-Noobj ~}\\\mbox{[71], RAP ~}\\\mbox{(IPatch) [89] ~} \\\normalsize \vspace{2mm}\textbf{in 2021 ~\hspace{5mm}} }}; %
  % \node[text width=2.6cm] at ([yshift=-10pt]mat-6-5) {\cbox{\small \mbox{DiAP [58],}\\\mbox{Patch-Noobj [71],}\\\mbox{RAP (IPatch) [89]} \\\normalsize \vspace{2mm}\textbf{in 2021 ~} }}; % \\\mbox{}
  % \node at ([yshift=-10pt, xshift=10pt]mat-6-5) {\cbox{\small \mbox{DiAP [58], ~}\\\mbox{Patch-Noobj ~}\\\mbox{[71], RAP ~}\\\mbox{(IPatch) [89] ~} \\\normalsize \vspace{2mm}\textbf{in 2021 ~\hspace{5mm}} }}; %
  
  % \iffalse%
  \node at ([yshift=-10pt]mat-6-1) {\cbox{\small \mbox{~}\vspace{2.mm} \\\mbox{\cite{brown2017adversarial}} \\\mbox{~} \\\normalsize \vspace{2mm}\textbf{in 2017} }}; % [37]
  \node at ([yshift=-10pt]mat-6-2) {\cbox{\small \mbox{~}\vspace{2mm} \\\mbox{DPatch \cite{liu2018dpatch},}\\\mbox{and \cite{song2018physical}} \\\normalsize \vspace{2mm}\textbf{in 2018} }}; % [59], [67]
  \node at ([yshift=-10pt]mat-6-3) {\cbox{\small \mbox{~}\vspace{2mm} \\\mbox{PS-GAN \cite{liu2019perceptual},}\\\mbox{and \cite{thys2019fooling}} \\\normalsize \vspace{2mm}\textbf{in 2019} }}; % [55], [69]
  \node at ([yshift=-10pt]mat-6-4) {\cbox{\small \mbox{~}\vspace{2.mm} \\\mbox{\cite{den2020adversarial, yamanaka2020adversarial}} \\\mbox{~} \\\normalsize \vspace{2mm}\textbf{in 2020} }}; % [70, 102]
  % \\\mbox{~}
  % \fi%
  
  \node[rotate = 90] at ([xshift=-54pt]mat-3-1.north)
    {\textsc{Illustrations}}; % xshift=-53pt, yshift=-19pt xshift=-0.5cm
  \node at ([yshift=-17pt, xshift=0.4cm]mat-8-3.south)
    {\textsc{Significant contribution in continuous years}};

  % Erase some visible lines outside the arrow
  \fill[white] (mat-1-6.north east) -- (mat-5-7.north east)
    -- (mat-1-7.north east) -- cycle;
  \fill[white] (mat-8-6.north east) -- (mat-5-7.north east)
    -- (mat-8-7.north east) -- cycle;

  % Draw the arrow tip
  \shade[top color=clrB!10, bottom color=clrB!10,
    middle color=clrB!70, draw=white, ultra thick] % !21, !80
  % \shade[top color=colfour!70, bottom color=colfour!70,
  %   middle color=colseven, draw=white, ultra thick] 
    (mat-1-6.north) -- (mat-5-7.north) -- (mat-8-6.south) -- 
    (mat-8-6.south east) -- (mat-5-7.north east) -- (mat-8-6.south east) -- 
    (mat-5-7.north east) -- (mat-1-6.north east) -- cycle;

  % The slanted "Margin" labels
  \begin{scope}[decoration={markings,
    mark=at position .5 with \node[transform shape] {\small \textsf{Adversarial patches}};}] % Adversarial attacks
  %  mark=at position .5 with \node[transform shape] {\small\bf Adversarial patches};}] % Adversarial attacks
    
  \path[postaction={decorate}] 
    ( $ (mat-1-6.north)!0.5!(mat-1-6.north east) $ )
    -- ( $ (mat-5-7.north)!0.5!(mat-5-7.north east) $ );
  \path[postaction={decorate}] 
    ( $ (mat-5-7.north)!0.5!(mat-5-7.north east) $ )
    -- ( $ (mat-8-6.south)!0.5!(mat-8-6.south east) $ );
  \end{scope}

  % The braces
  \draw[decorate, decoration={brace, mirror, raise=6pt}]
    (mat-1-1.north west) -- (mat-5-1.north west);
  \draw[decorate, decoration={brace, mirror, raise=6pt}]
    (mat-8-1.south west) -- (mat-8-6.south);
\end{tikzpicture}
% \end{minipage}
% \fi% }
\vspace{-.9em}% global scale = .88, .94
\caption{Important advances in adversarial patch attacks. 
Note that we only highlight some of adversarial attacks here, however there are more.
% We only highlight some of adversarial attacks here and there are more. % of them
}\label{fig:timeline}
\end{figure*}
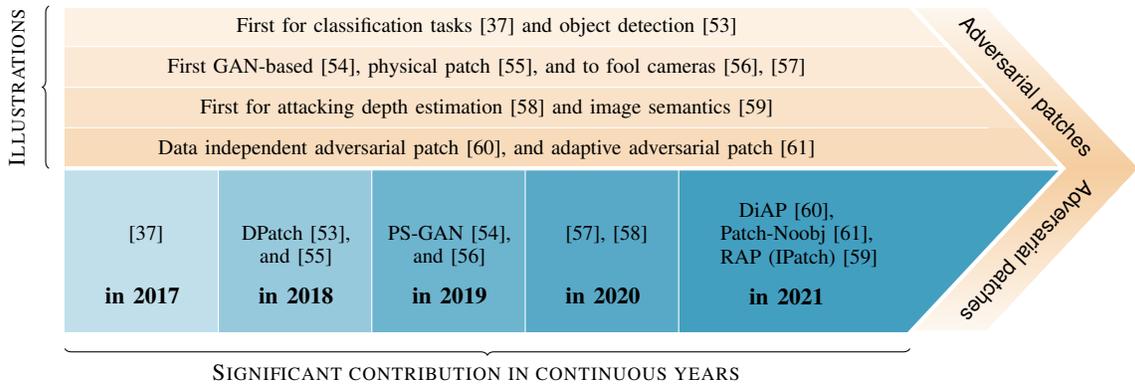

\subsection{Patch attacks for classification tasks}

% realisable

The concept of \textit{adversarial patch} attacks that are generated through training was firstly demonstrated by Brown \etal{} \cite{brown2017adversarial} in 2017, presenting a universal and targeted attack on real-world physical objects. 
Although primitive, this attack was also highly robust and practical, building a foundation for subsequent adversarial patches. 
Brown \etal{} \cite{brown2017adversarial} also presented camouflaged patches under constraints to force the similarity between the final patch and the starting patch. 
This attack is usually created in a relatively larger size and consequently is more evident to human eyes. 
Brown \etal{} \cite{brown2017adversarial} claimed that the region with the patch posing a security concern would become the most salient feature in the image, which was invalidated and thus distinguished it from LaVAN \cite{karmon2018lavan}. 
Karmon \etal{} \cite{karmon2018lavan} proposed \textit{LaVAN (Localised and Visible Adversarial Noise)} in a smaller size than adversarial patches \cite{brown2017adversarial} with similar attack effectiveness, focusing more on the model weaknesses that led to misclassification. 
LaVAN was the first to introduce some stealth properties in patch attacks. 
Karmon \etal{} \cite{karmon2018lavan} conducted patch perturbations in both images and networks for which the latter was more effective though not physically attainable. 
While it was more effective, LaVAN lacked robustness across transformations (like rotation) and locations over the image. 
Another attempt to make the adversarial patch less suspicious to human eyes was the \textit{adversarial QR patch} \cite{chindaudom2020adversarialqr}, \cite{chindaudom2022surreptitious}. 
It was created using a masked patch initialized with a QR pattern and trained later to make successful attacks. 
% Chindaudom claimed that QR patches had higher losses than their counterparts \cite{brown2017adversarial,karmon2018lavan}. 
QR patches opened the door to a new dimension where adversarial patches were trained using QR codes or some other strict patterns to mislead the human intuition. 
However, the harder one tried to obfuscate the patch from human detection, the less effective these attacks were. 
As we can see from the three aforementioned concepts, patch attacks are effective yet noticeable due to their irregular structure and unnatural appearance in the scene. Most efforts that are taken to reduce the identification of patches are observed to lead to the sacrifice of attacking behaviours. 
The reasons behind this stem from the fact that producing stronger attacks requires large perturbations without many structural constraints like in the QR patch \cite{chindaudom2020adversarialqr} and that the lack of constraints on perturbations leads to irregular and random patterns in the perturbation learning. 
Hence there exists a trade-off between patch identification and attack effectiveness. 
However, Liu \etal{} \cite{liu2019perceptual} argued that understanding the network perceptual sensitivity to adversarial patches could help to design more visually natural patches with strong attacking capability, which had been overlooked by existing patch attacks by then. 
They designed \textit{PS-GAN (Perceptual-Sensitive Generative Adversarial Networks)} to improve the visual fidelity and enhance the attacking ability, which was also demonstrated to have good transferability across network structures. 
\iffalse%
To prove this, they designed PS-GAN (perceptual-sensitive generative adversarial networks) to improve the visual fidelity and enhance the attacking ability. 
In this technique, the generator served to generate similar patches as the input seed patch, restricting the generated patches from having a suspicious pattern, while the discriminator helped to bring perceptual harmony between the attacked image and the original image, achieving visual fidelity. 
Moreover, an attention mechanism is also adopted to locate the most critical area in the scene, producing attacks with the maximal efficiency. 
\fi%
\iffalse%
PS-GAN was demonstrated to have a good transferability across network structures. 
Besides, choosing different seed patches would not cause a negative effect on adversarial patches applied by PS-GAN. 
\fi%
%%
%%
% namely a local-patch-based attack. 
Alternatively, Gittings \etal{} \cite{gittings2019robust} proposed an image reconstruction technique using deep image prior (DIP) \cite{ulyanov2018deep} to develop imperceptible perturbations that were robust to affine deformations, namely an attack based on local patches. 
They also claimed that the proposed image reconstruction helped achieve greater flexibility for perturbations across the whole image.

Unlike these five patches that were trained on the training data for deep neural networks, Zhou \etal{} \cite{zhou2021data} proposed \textit{DiAP (Data-independent Adversarial Patch)} to fool the target model without any knowledge of the training data. 
In this technique, non-targeted attacks were generated by optimising a spurious activation objective to deceive the features learned on each layer in the model and then were transformed into targeted ones by extracting important features from the background of the target class. 
However, the evaluation of DiAP on the whitebox-single model and whitebox-ensemble did not show superiority over the original adversarial patch \cite{zhou2021data}.

\subsection{Patch attacks for detection tasks}

The transferability of adversarial patches trained for classification usually fails to apply to object detection because of the inherent differences between their targets \cite{liu2018dpatch}. 
The model trained for detection tasks is expected to identify the location and then classify the object in a given proposed region, while in classification, only one object in the image is expected to be classified correctly. 
Adversarial patches for classification tasks perform an attack by training a patch to generate more salient features than those present in the image, thereby misleading the CNN models \cite{brown2017adversarial}. 
However, in typical detection tasks, well-known architectures (such as YOLO \cite{redmon2016you} or R-CNN \cite{girshick2014rich,girshick2015fast,ren2015faster}) involve generating many region proposals to initially locate objects and then classify them. 
Therefore, instead of one single salient feature in an image, several proposed objects and their bounding boxes need to be attacked in objection detection scenarios.

% which inspired Lee and Kolter \cite{lee2019physical}, who proposed 
After Brown \etal{} \cite{brown2017adversarial} demonstrated the effectiveness of adversarial patches in classification, Liu \etal{} \cite{liu2018dpatch} extended their idea to object detection, called \textit{DPatch (adversarial patch attack on object detectors)}, where the patch was iteratively trained to attack bounding box regression and object classification concurrently. 
DPatch could perform both untargeted and targeted attacks depending on how the patch was trained and empirically depicted decent attack transferability across datasets. 
However, DPatch was restricted to digital image scenarios, which inspired Lee and Kolter \cite{lee2019physical} to propose subtle modifications to make it more powerful and extend the ability to mislead object detection in real-world applications. 
Lee and Kolter \cite{lee2019physical} also claimed their patch attack was invariant to different lighting conditions, locations, or transformations even in real-world scenes. 
Moreover, DPatch contained a fundamental flaw where the pattern produced in the patch was not clipped in the allowable range for RGB pixels and thereby produced no actual image \cite{lee2019physical}. 
% allowance
Therefore, % Lee  \etal{} \cite{lee2019physical}  proposed Madry \etal{} \cite{madry2018towards} 's 
Madry \etal{} \cite{madry2018towards} proposed \textit{PGD (Projected Gradient Descent) technique} where the projection operation restricted the values of pixels within the permitted boundary (\ie{} RGB range), allowing for transferability of attacks to the real-world. 
However, the effectiveness of these patches was reduced as the objects were farther away from them. In this case, increasing the size of patches led to a stronger adversary. 
Afterwards, Wu \etal{} \cite{wu2020dpattack} proposed \textit{DPAttack (Diffused Patch Attacks)} which aimed to perturb a small image area and effectively attack many features in the scene. 
% Although quite effective it was across model architectures between YOLO and R-CNN, the patch did not account for much transferability to the physical world. 
% Wu's work
Then Huang \etal{} \cite{huang2021rpattack} extended their work \cite{wu2020dpattack} and proposed \textit{RPAttack (Refined Patch Attack)} by perturbing less pixels to create attacks as imperceptible as possible meanwhile retaining the effectiveness. 
They utilised the knowledge of key pixels to refine the patch and removed pixels in the patch that less affected the attack to reduce the number of unnecessary perturbations. 
Moreover, ensemble learning was carried out during the training phase in RPAttack to make the patch robust across model architectures.

The first adversarial patch in the physical world for object detection was demonstrated by Song \etal{} \cite{song2018physical}, inspired by $\mathrm{RP}_2$ (Robust Physical Perturbations) \cite{eykholt2018robust} for object classification. 
They presented the vulnerability of object detectors in the physical world and two types of attacks \ie{} the disappearance attack and the creation attack with stop signs on the road. 
In the former, the untargeted attack was performed by suppressing the generation of bounding boxes and thereby stopping the detector from detecting objects in the scene, while in the latter, the detector was fooled to detect non-existent objects in the scene. 
Despite the effectiveness of adversarial patches in both digital and real-world applications, most of the attacks discussed until now highlighted the feasibility in the scenes where intra-class variation was minimal. 

% fool models in automated surveillance cameras
Thys \etal{} \cite{thys2019fooling} attempted to fool models in surveillance cameras to detect persons, which were usually used in high-security zones to detect trespassing into a restricted area. 
This was challenging because the attack needed to be equally effective against people in front of the camera with different colors, sizes, clothing, orientations, and poses. 
Moreover, people were much more varied in terms of shapes and appearances than road signs which were usually consistent. For example, a stop sign appeared the same regardless of the environments it was in. 
% Moreover, people might appear in different contexts, unlike road signs. 
Thys \etal{} \cite{thys2019fooling} proposed a loss objective to reduce instead of the classification score, bringing attacks that performed exceptionally well in practice, although they required strong conditions on location and lacked transferability across model architectures as well. 
Another application of fooling cameras was the camouflage of military assets against aerial detection, which was an extended work of \cite{thys2019fooling} proposed by Hollander \etal{} \cite{den2020adversarial}. 
They modified the loss function slightly to make the patch more difficult for human eyes to detect. 
Hollander \etal{} \cite{den2020adversarial} empirically presented a trade-off between size and performance, where the attack was most effective with the large patch exactly over the asset/object of interest yet small patches (even including less colorful ones) on the object gave a better performance, depicting how vital the patch location was in the attack. 
As illegal drone usage increased for surveillance near military or defended areas, the patch camouflage application became of utmost importance and a potentially viable threat. 
On the other hand, it helped eradicate the requirement of manual camouflage nets to cover assets, because bigger assets or larger numbers of them often made it infeasible. 
Moreover, Lu \etal{} \cite{lu2021scale} proposed \textit{Patch-Noobj} to adaptively scale the patch size based on the size of attacked aircraft, showing attack transferability across both models and datasets. 

Stationary patches that were static in the scene \cite{liu2018dpatch,song2018physical,lee2019physical,thys2019fooling,den2020adversarial} would become less effective if the relative position of the camera changed with respect to an attacked image, like a moving car on the road. 
The patch performance became uncertain due to two reasons: %\\
1) the camera angle changed by relative motion; %\\
2) the change of the camera's field of view led to different sizes of target objects in each frame. %\\
Most of these patches worked for planar objects\footnote{%
Planar is with respect to the object's real-world geometry. For example, road signs are planar, but a human body is not. 
% Planar is with respect to the object's geometry, like road signs are planar but a human body is not. 
% Planar is with respect to its geometry, like road signs are planar but a human body is not. 
} except for \cite{thys2019fooling}, yet objects to the attacks in most real-world scenarios were non-planar. 
Therefore, Hoory \etal{} \cite{hoory2020dynamic} designed the \textit{Dynamic Adversarial Patch} to be invariant to the camera's position by switching between the trained patches to make the attack dynamic to the upcoming scene. 
% the scene at hand. 
Moreover, multiple screens were placed at different locations to attack detection when the camera's viewpoint changed or several cameras were present at the scene. 
Semantic adversary features were also introduced to prevent semantically related classes (such as car, bus, or truck) from having the same influence on autonomous driving scenarios. 
Note that the dynamic patch \cite{hoory2020dynamic} was the first attempt to make an adversary adaptable to dissimilar situations. 
However, drawbacks also existed that required future explorations of real-world scenarios, including: 
transferability across architectures, environments, or models, as well as cost requirements for screens or LEDs.

Additionally, Zolfi \etal{} \cite{zolfi2021translucent} proposed the \textit{translucent patch} to apply it on the camera lenses, while most attacks were proposed to attack objects. 
This patch used for object detection was able to attack the target object only and leave the rest of the objects in the scene untouched, which was more challenging than that of \cite{li2019adversarial} for object classification. 
Zolfi \etal{} diligently crafted the patch to incorporate necessary features in the attack including the patch structure, region-level patch blending, shape positioning, and shearing. 
They also presented empirically that the attack was transferable across model architectures like R-CNN. 
Wang \etal{} \cite{wang2021towards} designed the \textit{invisibility patch} to attack target classes only in the scene by making the target object invisible to the detector. 
The patch is trained iteratively to minimise the loss of a detection score concerning the target class. 
% The beauty of this attack lies in its high transferability across datasets, architectures, and from digital to physical world. 
% Unlike other attacks, 
This attack is highly transferable across datasets, architectures, and from the digital to the physical world. 
Wang \etal{} \cite{wang2021towards} proposed to display patches on a portable screen instead of a poster or sticker, showing a good performance in the physical world. 
However, the limitation of this attack existed as the patch was required to be present exactly over the target image and the need of a hardware screen also increased the cost to perform an attack. 
%%
% proposed an \textit{attention-guided digital adversarial patch}
In the context of most existing attacks using random gradient descent to generate and adjust the patch, Lang \etal{} \cite{lang2021attention} proposed \emph{AGAP (Attention-Guided digital Adversarial Patch)} using high feature density regions in the image to calculate the location and size of the generated patch. 
A heat map was used to identify the important features of the image and help locate the object position. 
However, the heat map was constructed on digital images, which caused the feature density to vary significantly in real-world examples and resulted in reduced attack success rate.

%

\begin{table*}[t]%
\centering\caption{%
A brief summary of various adversarial patch attacks for different vision tasks. 
Note that the last column called ``Physical Demonstration'' represents whether the authors presented the transferability of their patch attack(s) into the physical world. 
}\label{tab:comparison}
\vspace{-1.5em}\renewcommand\tabcolsep{.46em}% 1.7pt
\subfloat[Brief summary for \textsc{Classification}]{
\scalebox{.97}{% .1
% \begin{tabular}{r| C{.46\textwidth} cc}
\begin{tabular}{r| C{.4\textwidth} cc}% .39
  \toprule
  \multicolumn{1}{c|}{\bf Patch Attack} & {\centering\bf Attacked Model Architecture(s)} & {\centering\bf Dataset(s)} & \tabincell{c}{\textbf{Physical}\\ \textbf{Demonstration}} \\
  %%%%
  \midrule%
  Adversarial patch \cite{brown2017adversarial} & VGG16, VGG19, ResNet50, InceptionV3, Xception & ImageNet & Yes \\
  LaVAN \cite{karmon2018lavan} & InceptionV3 \cite{szegedy2016rethinking} & ImageNet & No \\
  QP patch \cite{chindaudom2020adversarialqr,chindaudom2022surreptitious} & InceptionV3 \cite{szegedy2016rethinking} & ImageNet \cite{russakovsky2015imagenet} & No \\
  PS-GAN$^a$ \cite{liu2019perceptual} & VGG16, ResNet34, VY \cite{yadav2016ptwo}, and their variants & ImageNet \cite{deng2009imagenet}, GTSRB \cite{houben2013detection} & Yes \\
  Local patch via DIP$^b$ \cite{gittings2019robust} & VGG19, InceptionV3 & ImageNet \cite{deng2009imagenet} & No \\
  DiAP \cite{zhou2021data} & VGG16, VGG19 \cite{simonyan2014very}, ResNet50 \cite{he2016deep}, InceptionV3 \cite{szegedy2016rethinking}, Xception \cite{chollet2017xception} & ImageNet & Yes \\
  Camera stickers$^c$ \cite{li2019adversarial} & ResNet50 \cite{he2016deep} & ImageNet \cite{deng2009imagenet} & Yes \\
  \bottomrule
  %%%%
  \multicolumn{4}{l}{\scriptsize$^a$Adversarial patches were generated by PS-GAN where Liu \etal{} \cite{liu2019perceptual} employed U-Net with customisation on QuickDraw \cite{jongejan2017quick} correspondingly.}\\%
  \multicolumn{4}{l}{\scriptsize$^b$Gittings \etal{} \cite{gittings2019robust} adapted the local patch based attack to generate adversarial images via DIP (deep image prior) \cite{ulyanov2018deep} reconstruction.}\\%
  \multicolumn{4}{l}{\scriptsize$^c$Li \etal{} \cite{li2019adversarial} proposed a physical camera-based attack for object classification.}%
\end{tabular}
}}
\\ \vspace{-.51em}
\subfloat[Brief summary for \textsc{Detection}]{
\scalebox{.94}{% .97
% \begin{tabular}{ R{.24\textwidth}| C{.32\textwidth} C{.3\textwidth} c}
\begin{tabular}{ R{.275\textwidth}| C{.28\textwidth} C{.25\textwidth} c}% .24-.27-.24 % .27-.28-.25
  \toprule
  \multicolumn{1}{c|}{\bf Patch Attack} & {\bf Attacked Model Architecture(s)} & {\bf Dataset(s)} & \tabincell{c}{\bf Physical\\ \bf Demonstration} \\
  \midrule
  DPatch \cite{liu2018dpatch} & YOLOv2 \cite{redmon2017yolo9000}, Faster R-CNN \cite{ren2015faster} & Pascal VOC 2007 \cite{everingham2008pascal} & \hspace{.1em} No$^d$ \\
  Extended $\mathrm{RP}_2 \,^e$ \cite{song2018physical} & YOLOv2 \cite{redmon2017yolo9000}, Faster R-CNN \cite{ren2015faster} & Seld-made data (indoors and outdoors) & Yes \\
  Physical patch +PGD$^e$ \cite{lee2019physical} & YOLOv3 \cite{redmon2018yolov3} & MS COCO 2014 \cite{lin2014microsoft} & Yes \\
  Patch against person detectors \cite{thys2019fooling} & YOLOv2 \cite{redmon2017yolo9000} & INRIA \cite{dalal2005histograms}\,$^{e}$ & Yes \\% (OBJ-CLS)
  % Patch against aerial detection \cite{den2020adversarial} & YOLOv2 & DOTA \cite{xia2018dota} & No \\
  DPAttack \cite{wu2020dpattack} & YOLOv4 \cite{bochkovskiy2020yolov4}, Faster R-CNN \cite{ren2015faster} & Alibaba$^{f}$ & No \\
  RPAttack \cite{huang2021rpattack} & YOLOv4 \cite{bochkovskiy2020yolov4}, Faster R-CNN \cite{ren2017faster} & Alibaba$^{f}$, Pascal VOC 2007 \cite{everingham2010pascal} & No \\
  Patch against aerial detection \cite{den2020adversarial} & YOLOv2 & DOTA \cite{xia2018dota} & No \\
  Patch-Noobj \cite{lu2021scale} & {YOLOv3 \cite{redmon2018yolov3}, YOLOv5 \cite{jocher2020yolov5}, Faster R-CNN \cite{ren2015faster}} & {DOTA \cite{xia2018dota}, NWPU VHR-10 \cite{cheng2016learning}, RSOD \cite{xiao2015elliptic} } & \hspace{.4em} No\,$^{g}$ \\
  Translucent patch \cite{zolfi2021translucent} & {YOLOv2 \cite{redmon2017yolo9000}, YOLOv5 \cite{jocher2020yolov5}, Faster R-CNN \cite{ren2015faster}} & LISA \cite{mogelmose2012vision}, MTSD \cite{ertler2020mapillary}, BDD \cite{yu2020bdd100k} & \hspace{.4em} Yes\,$^{g}$ \\
  Dynamic patch \cite{hoory2020dynamic} & YOLOv2 \cite{redmon2017yolo9000} & Self-made data (annotated manually) & Yes \\
  Invisibility patch \cite{wang2021towards} & YOLOv3 \cite{redmon2018yolov3}, Faster R-CNN \cite{ren2015faster} & MS COCO 2017 \cite{lin2014microsoft}, Pascal VOC 2007 \cite{everingham2010pascal,everingham2015pascal} & \hspace{.4em} Yes\,$^{g}$ \\
  Patch exploiting contextual reasoning \cite{saha2019adversarial} & YOLOv2 & Pascal VOC 2007, KITTI & No \\
  AGAP (Attention-guided patch) \cite{lang2021attention} & YOLOv2 \cite{redmon2017yolo9000}, Faster R-CNN \cite{ren2017faster} & MS COCO 2017 & \hspace{.4em} No\,$^{g}$ \\
  \bottomrule
  %%%%
  % L{.91\textwidth} + L{1.\textwidth}x2 + L{.98\textwidth}
  \multicolumn{4}{L{.91\textwidth}}{\scriptsize$^d$Liu \etal{} \cite{liu2018dpatch} tested the transferability of DPatch trained by MS COCO on Pascal VOC, where VGG16 and ResNet101 served as Faster R-CNN's basic networks.}\\
  \multicolumn{4}{L{.98\textwidth}}{\scriptsize$^e$Eykholt \etal{} \cite{song2018physical} modified $\mathrm{RP}_2$ (robust physical perturbations) \cite{evtimov2017robust} and experimented upon self-recorded videos (in a mix of lab settings). Lee and Kolter \cite{lee2019physical} created adversarial patches based on untargeted PGD (projected gradient descent) \cite{madry2018towards} with expectation over transformation. Thys \etal{} \cite{thys2019fooling} trained the YOLO detector on MS COCO 2014 and only tested on INRIA, because the variety that COCO \cite{lin2014microsoft} and Pascal VOC \cite{everingham2010pascal} contain made it difficult to put their patch in a consistent position. }\\
  \multicolumn{4}{L{.97\textwidth}}{\scriptsize$^f$Ren \etal{} \cite{ren2015faster} tested DPAttack using data from Alibaba Tianchi competition and Alibaba-Tsinghua adversarial challenge on object detection. Later, Ren \etal{} \cite{ren2017faster} tested RPAttack using data from Alibaba-Tsinghua adversarial challenge on object detection that sampled 1,000 images from MS COCO 2017 test set \cite{lin2014microsoft}. }\\
  \multicolumn{4}{L{.98\textwidth}}{\scriptsize$^g$% .91-.96-.94-.98
  Lu \etal{} \cite{lu2021scale} empirically demonstrated Patch-Noobj's transferability in three scenarios: dataset-to-dataset, model-to-model, and joint both. Wang \etal{} \cite{wang2021towards} also demonstrated the transferability of the invisibility patch across models and datasets. %
  % }\\ \multicolumn{4}{L{1.\textwidth}}{\scriptsize$^h$%
  Zolfi \etal{} \cite{zolfi2021translucent} demonstrated the transferability of the translucent patch attack on two attacked models: YOLOv2 \cite{redmon2017yolo9000} and Faster R-CNN \cite{ren2015faster}. Lang \etal{} \cite{lang2021attention} tested the transferability of AGAP (attention-guided adversarial patch) between YOLO and Faster-RCNN.}
\end{tabular}
}}
\end{table*}

\begin{table*}[t]
\centering\caption{%
A brief summary of various adversarial patch attacks for different vision tasks (cont)%. 
: \textrm{(c) Brief summary for \textsc{Other visual tasks}}. 
Note that the last column called ``Physical Demonstration'' represents the same meaning as that in Table~\ref{tab:comparison}.
}\label{tab:compromise}% (cont.)
\renewcommand\tabcolsep{.46em}%
\scalebox{.97}{% 1.
% \begin{tabular}{c|| R{.16\textwidth} | C{.27\textwidth} C{.2\textwidth} c}%
\begin{tabular}{c|| R{.165\textwidth} | C{.285\textwidth} C{.2\textwidth} c}% .165-.285-.205
  \toprule
  {\bf Vision Task} & \multicolumn{1}{c|}{\bf Patch Attack} & {\bf Attacked Model Architecture(s)} & {\bf Dataset(s)} & \tabincell{c}{\bf Physical\\ \bf Demonstration} \\
  \midrule
  \textsc{Object recognition}
  & IPatch\,$^{i}$ \cite{mirsky2021ipatch} & YOLOv3 \cite{redmon2018yolov3} & CamVid \cite{brostow2009semantic} & No \\
  \textsc{Segmentation} 
  & IPatch (\aka{} RAP) \cite{mirsky2021ipatch} & Unet++ \cite{zhou2018unet++}, Linknet \cite{chaurasia2017linknet}, FPN \cite{lin2017feature}, PSPNet \cite{zhao2017pyramid}, PAN \cite{li2018pyramid}\,$^h$ & CamVid \cite{brostow2009semantic} & No \\
  & Scene-specific patch \cite{nesti2022evaluating} & DDRNet \cite{hong2021deep}, BiSeNet \cite{yu2018bisenet}, ICNet \cite{zhao2018icnet}, PSPNet \cite{zhao2017pyramid} & CityScapes \cite{cordts2016cityscapes}, CARLA & Yes \\
  \midrule
  \textsc{Depth Estimation} 
  & MonoDepth$^k$ \cite{yamanaka2020adversarial} & Stereo*$\to$Mono \cite{guo2018learning}, MSLPG \cite{lee2019big}\,$^k$ & KITTI \cite{uhrig2017sparsity,geiger2013vision} & Yes \\%
  \midrule
  \textsc{Optical Flow}
  & Patch against optical flow networks \cite{ranjan2019attacking} & FlowNetC \cite{dosovitskiy2015flownet}, FlowNet2 \cite{ilg2017flownet}, SpyNet \cite{ranjan2017optical}, PWC-Net \cite{sun2018pwc}, Back2Future \cite{janai2018unsupervised}, LDOF \cite{brox2010large}, EpicFlow \cite{revaud2015epicflow} & KITTI \cite{geiger2012we} & Yes \\
  \midrule
  \textsc{Image retrieval}\,$^j$%
  & AP-GAN \cite{zhao2020ap} & VGG16, ResNet101 \cite{radenovic2018fine} & Oxford5K \cite{philbin2007object}, Paris6K \cite{philbin2008lost}, retrieval-SfM-30k \cite{radenovic2016cnn} & Yes \\
  \textsc{Person ReID}\,$^j$%
  & AP-GAN \cite{zhao2020ap} & ResNet50 \cite{zheng2017person}, MGN \cite{wang2018learning} & Market-1501 \cite{zheng2015scalable}, DukeMTMC-ReID \cite{zheng2017unlabeled} & No \\
  \textsc{Vehicle ReID}\,$^j$%
  & AP-GAN \cite{zhao2020ap} & open-VehicleReID \cite{chen2019region} & VeRi776 \cite{liu2016large} & No \\
  \bottomrule
  %%%%
  % L{.91\textwidth} x4
  \multicolumn{5}{L{.95\textwidth}}{\scriptsize$^h$Mirsky \cite{mirsky2021ipatch} employed 37 different models to train and attack actually that were combinations of eight encoders and five SOTA (state-of-the-art) segmentation architec- tures listed above. The encoders included vgg19, densenet121, efficientnet-b4, efficientnet-b7, mobilenet\_v2, resnext50\_32x4d, dpn68, and xception. }\\
  \multicolumn{5}{L{.95\textwidth}}{\scriptsize$^i$After analysing the patch against segmentation models, Mirsky \cite{mirsky2021ipatch} presented it as an extention using the same training framework on other semantic models. }\\
  \multicolumn{5}{L{.95\textwidth}}{\scriptsize$^j$Zhao \etal{} \cite{zhao2020ap} evaluated AP-GAN in three \textsc{Content-based image retrieval} systems, where a generator in the GAN-based framework was employed to generate a deceptive adversarial patch for each input image, perturbing effectively yet naturally and subtly.}\\
  \multicolumn{5}{L{.95\textwidth}}{\scriptsize$^{k}$%
  Yamanaka \etal{} \cite{yamanaka2020adversarial} proposed an adversarial patch attack for CNN-based monocular depth estimation methods. 
  Guo \etal{} \cite{guo2018learning} trained monocular depth networks by distilling cross-domain stereo networks including StereoNoFt, StereoUnsupFt, and StereoSupFt100. Lee \etal{} \cite{lee2019big} placed MSLPG (Multi-Scale Local Planar Guidance) layers that were located on multiple stages in the decoding phase to recover internal feature maps. %
  }
\end{tabular}
}
\end{table*}

\subsection{Patch attacks for other vision-based tasks}

Up to now, the adversarial patches that we discussed focus on either object classification or object detection applications. 
There are other adversarial patches as well. 
For example, Mirsky \cite{mirsky2021ipatch} proposed \textit{RAP (Remote Adversarial Patch, \aka{} IPatch)} to attack image semantics, which was simple to implement as the trained patch could be placed anywhere in the image. 
This attack became useful when the target object was impractical to reach (\eg{} sky or moving car) or applying the patch on the target became highly obvious (\eg{} sticker on a stop sign), because the patch could affect the target class or semantics even if it was located far away from the target object. 
\iffalse%
Like most of the adversarial patches, RAP was robust to attack the scene at different locations or scales. 
RAP was also demonstrated to be transferable for object detection tasks but with relatively lower success rate than patch attacks where adversarial stickers were directly applied over the object due to the trade-off between reliability and perception. 
Moreover, the flexibility and stealth characteristics of RAP made it more potent for some certain applications. 
\fi%
Moreover, Yamanaka \etal{} \cite{yamanaka2020adversarial} crafted a patch over the target region in an image to mislead the model in detecting its actual depth.  
They argued that monocular depth estimation was vulnerable with only one image because its inherent characteristic was based on non-depth features such as colors and vertical positions. 
This attack was one of the earliest works towards attacking depth estimation in the physical world and presented the transferability to it as well. 
Additionally, Zhao \etal{} \cite{zhao2020ap} proposed \textit{AP-GAN (Adversarial Patch-GAN)}, a technique that was similar to PS-GAN for crafting adversarial patches using generative adversarial networks (GANs). 
AP-GAN misled the identifier to return dissimilar images for input queries in image retrieval applications. 
This was a semi-whitebox attack because the target model was only required during training and not during inference. 
AP-GAN could be physically applied through stickers or posters and remained similarly effective under affine transformations in the physical world even when the relative position of objects and the camera did change. 
AP-GAN was also the first adversarial patch for image retrieval, showing good transferability across detector models and datasets.

% \section{Patch detection/defense}
\section{Patch detection/defence}
\label{sec: patch_defense}

Adversarial patch attacks have been demonstrated as potential and practical threats in real-world scenarios. 
Therefore, developing detection and defence against adversarial patches becomes increasingly important due to widespread use in security systems and safety-critical domains, such as robust autonomous vehicles on the road, CCTV camera and drone surveillance, and potential intruders along the perimeters of buildings or in military applications. 
Patch attacks are relatively easy to craft yet difficult to defend against. 
Designing defences for imperceptible norm-based attacks utilises an adversarial version of the original data distribution to help the model learn the behaviour of adversarial inputs. 
The failure of adversarial training would be instrumental in designing diverse defence techniques for patch attacks, such as using inherently robust architectures and saliency maps. 
The possibility of patch attacks in different forms has also inspired the formulation of certified defences. 
Defences for patch attacks are typically viewed as a detection problem. 
Once the patch's location is detected, the suspected region would be either masked or inpainted to mitigate the adversarial influence on the image. 
In this section, we discuss the state-of-the-art defence techniques against patch attacks to produce robust vision models. 
The defences presented here have been grouped into logical categories and arranged in chronological order for ease of understanding.

% \subsection{Defenses based on the saliency map}
\subsection{Defences based on saliency maps}
\label{subsec:dbsm}

Adversarial patches are observed to increase high-frequency perturbations in local regions of an image and generally do not overlap with salient objects. 
In localised attacks, dense clustering around perturbations is often observed in saliency, while in contrast, the output without the attack is affected by the entire image rather than a small local area, which is helpful to detect the location of patch attacks. 
The idea of saliency-based defences is simple and attractive to detect adversarial patches. 
However, its performance usually greatly depends on the quality of saliency maps, because the intended objects might be highly active in the saliency map sometimes.

% firstly proposed the defense against
For instance, 
Hayes \cite{hayes2018visible} first proposed \emph{DW (Digital Watermarking)}, a defence against adversarial patches for non-blind and blind image inpainting, inspired by the procedure of digital watermarking removal. 
A saliency map of the image was constructed to help remove small holes and mask the adversarial image, blocking adversarial perturbations. 
This was an empirical defence with no guarantee against adaptive adversaries. 
It also laid the foundation for the following certified defences of patch attacks. 
Naseer \etal{} \cite{naseer2019local} proposed \textit{LGS (Local Gradient Smoothing)} to suppress highly activated and perturbed regions in the image without affecting salient objects. 
Specifically, the irregular gradients were regularised in the image before being passed to a deep neural network (DNN) model for inference. 
LGS could achieve robustness with a minimal drop in clean accuracy because it was based on local region processing in contrast to the global processing on the whole image as done by its counterparts. 
Chou \etal{} \cite{chou2020sentinet} proposed \textit{SentiNet} for localised universal attacks to use the particular behaviour of adversarial misclassification to detect an attack, which was the first architecture that did not require prior knowledge of trained models or adversarial patches. 
Salient regions were used to help observe the model's behaviour. 
SentiNet was demonstrated to be empirically robust and effective even in real-world scenarios. 
However, it evaluated adversarial regions by subtracting the suspicious region, which might at times cause false adversarial region proposals. 
Moreover, the suspicious adversarial region was placed at random locations in the preserving image, which possibly occluded the main objects in the scene resulting in incorrect predictions. 

It is worth mentioning that none of these three empirical defences \cite{hayes2018visible,naseer2019local,chou2020sentinet} was effective enough to mitigate general patch attacks. 
To this end, Chen \etal{} \cite{chen2021turning} proposed \textit{Jujutsu} to detect and mitigate robust and universal adversarial patch attacks by leveraging the attacks' localised nature via image inpainting. 
A modified saliency map \cite{smilkov2017smoothgrad} was used to detect the presence of highly active perturbed regions, which helped to place suspicious extracted regions in the least salient regions of the preserved image and avoid occlusion with main objects in the image. 
Jujutsu showed a better performance than other empirical defences in terms of both robust accuracy and low false-positive rate (FPR), across datasets, patches of various shapes, and attacks that targeted different classes.

% \subsection{Defenses based on the adversarial training}
\subsection{Defences based on adversarial training}
\label{subsec:dbat}

The conventional solution of adversarial training is not achievable for patch attacks, because the combination of the possible patch placement on an image is numerous and the task of producing new patches at each iteration is infeasible. 
Patch attacks are local and perceptible, implying a highly active and overt region in the image. 
Besides, the magnitude of noise in patch attacks is relatively much higher, unlike norm-based covert noise. 
Adversarial training with patch attacks will lead to training on a divergent data distribution from the original one, impacting the decision boundary and sacrificing natural accuracy. 
Hence designing defences based on adversarial training requires some careful consideration and modification of either the training process or the architecture level. 
In this subsection, we discuss the defences built on modified versions of adversarial training.

% \textit{DC-GAN (deep convolutional generative adversarial network)} 
% DC-GAN was % employed GAN 
% They proposed \emph{VaN (Vax-a-Net)}, which adapted DC-GAN (deep convolutional generative adversarial network) \cite{radford2015unsupervised} to train a pretrained CNN model for adversarial patches. 
% VaN was demonstrated transferability across traditional CNN model architectures and lower inference time than LGS \cite{naseer2019local} and DW \cite{hayes2018visible} defenses.
Gittings \etal{} \cite{gittings2020vax} were the first to propose training-time defence against patch attacks, while two defences \cite{hayes2018visible,naseer2019local} that exploited highly active visual behaviours on saliency maps were also based at inference time. 
They proposed \emph{VaN (Vax-a-Net)} to defend against adversarial patch attacks in image classification, where DC-GAN (Deep Convolutional Generative Adversarial Network) \cite{radford2015unsupervised} was adapted to synthesise effective adversarial patches
and train the model simultaneously to defend against those patches. 
% \emph{D-VaN (Vax-a-Net defence)}
% VaN was demonstrated transferability across traditional CNN model architectures and lower inference time than LGS \cite{naseer2019local} and DW \cite{hayes2018visible} defenses. 
%%
In contrast to the model behaviours against imperceptible attacks, Rao \etal{} \cite{rao2020adversarial} developed an adversarial training technique to improve the model robustness against adversarial patches without sacrificing clean accuracy. 
They also ameliorated LaVAN \cite{karmon2018lavan} with additional features of location optimisation, bringing extra computational cost due to the full exploration of the space of all possible locations.

\begin{table*}[ht]
\centering
\caption{A brief summary of various adversarial patch detection/defences. 
Note that the last column called ``Physical Demonstration'' represents whether the authors presented empirical results of their patch defence(s) in the physical world. 
}\label{tab:defences}
\vspace{-1.5em}\renewcommand\tabcolsep{.46em}%
\subfloat[Defences based on \textsc{Saliency map}, \textsc{Adversarial training}, or \textsc{Small receptive fields}]{%
\scalebox{.97}{% .94
\begin{tabular}{r| C{.235\textwidth} C{.235\textwidth} C{.23\textwidth} c}
  \toprule
  \multicolumn{1}{c|}{\bf Patch Defence} % \tabincell{c}{\bf Patch Defence} 
  & \textbf{Attacked Model Architecture(s)} & \textbf{Against Attack(s)} & \textbf{Dataset(s)} & \tabincell{c}{\bf Physical\\\bf Demonstration} \\
  \midrule
  %%
  % Saliency map
  DW \cite{hayes2018visible} & VGG19, ResNet101, InceptionV3 & Adversarial patch \cite{brown2017adversarial}, LaVAN \cite{karmon2018lavan} & ImageNet\,$^l$ & No \\
  LGS \cite{naseer2019local} & InceptionV3 \cite{szegedy2016rethinking} & LaVAN \cite{karmon2018lavan} & ImageNet\,$^l$ & No \\
  SentiNet \cite{chou2020sentinet} & VGG16 \cite{simonyan2014very}, Faster R-CNN \cite{ren2015faster} & Adversarial patch \cite{brown2017adversarial}, backdoors \cite{gu2017badnets}, trojan triggers \cite{liu2017trojaning} & ImageNet,$^l$ LISA \cite{mogelmose2014traffic}, LFW \cite{huang2008labeled} & Yes \\
  Jujutsu \cite{chen2021turning} & ResNet18, ResNet50 & Adversarial patch \cite{brown2017adversarial} & ImageNet \cite{deng2009imagenet}, ImageNette \cite{howard2021image}, CelebA \cite{liu2015deep}, Place365 \cite{zhou2017places} & Yes \\
  % \hline
  % Adversarial training
  VaN \cite{gittings2020vax} & VGG19 \cite{simonyan2014very}, InceptionV3 \cite{szegedy2016rethinking}, Inception-ResNet-v2 (IRN-v2) \cite{szegedy2017inception} & Adversarial patch \cite{brown2017adversarial}, local patch via DIP \cite{gittings2019robust} & ImageNet \cite{deng2009imagenet} & Yes \\
  % \hline
  % Small receptive fields 
  PG \cite{xiang2020patchguard,xiang2021patchguard} & DS-ResNet,$^m$ BagNet & Adversarial patch$^m$ & ImageNet \cite{deng2009imagenet}, ImageNette \cite{howard2021image}, CIFAR10 \cite{krizhevsky2009learning} & No \\
  PG++ \cite{xiang2021patchguard++} & BagNet33 \cite{brendel2019approximating} & Adversarial patch & ImageNet \cite{deng2009imagenet}, ImageNette \cite{howard2021image}, CIFAR10 \cite{krizhevsky2009learning} & No \\
  DG \cite{xiang2021detectorguard} & YOLOv4 \cite{bochkovskiy2020yolov4,wang2021scaled}, Faster R-CNN \cite{ren2015faster}, PCD$^n$ & Localised patch hiding attacks$^n$ & Pascal VOC 2007 \cite{everingham2010pascal}, MS COCO 2014 \cite{lin2014microsoft}, KITTI \cite{geiger2013vision} & No \\
  % \hline
  %%
  \bottomrule
  \multicolumn{5}{l}{\scriptsize$^l$Data in \cite{hayes2018visible,naseer2019local} was from the ImageNet validation set. Data in \cite{chou2020sentinet} was from the Imagenet test set \cite{deng2009imagenet}.}\\
  \multicolumn{5}{L{.99\textwidth}}{\scriptsize$^m$DS-ResNet (De-randomised Smoothed ResNet) \cite{levine2020randomized} was used to instantiate an ensemble approach. All analysed network structures included ResNet50, BagNet9, BagNet17, BagNet33, DS25-ResNet50, and DS4-ResNet18. The defence performance was analysed against a single square adversarial patch.}\\
  \multicolumn{5}{L{.99\textwidth}}{\scriptsize$^n$Xiang and Mittal \cite{xiang2021detectorguard} explained that BagNet33 \cite{brendel2019approximating} was used as objectness predictors and that PCD (Perfect Clean Detector) served as one of the base detectors along with YOLOv4 \cite{bochkovskiy2020yolov4,wang2021scaled} and Faster R-CNN \cite{ren2015faster} where PCD was a hypothetical object detector simulated with ground-truth annotations.}\\
\end{tabular}
}}\\
\subfloat[\textsc{Certified defences}, \textsc{Domain specific defences}, and other defence techniques]{
\scalebox{.97}{% .94
\begin{tabular}{R{.16\textwidth}| C{.215\textwidth} C{.235\textwidth} C{.205\textwidth} c}
  \toprule
  \multicolumn{1}{c|}{\bf Patch Defence} % \tabincell{c}{\bf Patch Defence} 
  & \textbf{Attacked Model Architecture(s)} & \textbf{Against Attack(s)} & \textbf{Dataset(s)} & \tabincell{c}{\bf Physical\\\bf Demonstration} \\
  \midrule
  % Certified defences 
  CertIBP \cite{chiang2020certified} & CNN, fully-connected network & Sparse attacks & MNIST, CIFAR10 & No \\% Convolutional network
  PixelDP \cite{lecuyer2019certified} & DNNs (such as InceptionV3) & $\mathrm{L}_2$-norm attacks & MNIST, CIFAR10, CIFAR100, SVHN, ImageNet & No \\
  RCD \cite{lin2021certified} & ResNet9,$^o$ ResNet34 & Adversarial patch$^o$ & CIFAR10, ImageNet & No \\
  BagCert \cite{metzen2021efficient} & ResNet \cite{he2016deep} & Square patches & CIFAR10 \cite{krizhevsky2009learning}, ImageNet \cite{russakovsky2015imagenet} & No \\
  % \hline
  % Domain specific defences
  HyperNeuron \cite{co2021real} & ResNet50 \cite{he2016deep} & Adversarial patch \cite{brown2017adversarial} & ImageNet & Yes \\
  Defence for vehicle control \cite{pavlitskaya2020feasibility} & DriveNet \cite{hubschneider2017adding} & JSMA-based patches \cite{papernot2016limitations}, PGD-based patches \cite{madry2018towards} & CARLA simulator \cite{dosovitskiy2017carla} & No \\
  Detection for semantic segmentation (SS) \cite{nesti2022evaluating} & DDRNet \cite{hong2021deep}, BiSeNet \cite{yu2018bisenet}, ICNet \cite{zhao2018icnet}, PSPNet \cite{zhao2017pyramid} & EOT-based patches \cite{brown2017adversarial,xu2020adversarial}, scene-specific patches \cite{nesti2022evaluating} & CARLA simulator \cite{dosovitskiy2017carla}, Cityscapes \cite{cordts2016cityscapes} & Yes \\
  % \hline
  % Other defence techniques
  MR \cite{mccoyd2020minority} & SimpNet \cite{hasanpour2018towards}, VGG16 \cite{simonyan2014very}, Deotte model \cite{deotte2018how} & Adversarial patch \cite{brown2017adversarial} & MNIST, Fashion-MNIST, CIFAR10 & No \\% [Deo18]
  % MR \cite{mccoyd2020minority} & Non-indicated & Adversarial patch \cite{brown2017adversarial} & MNIST, Fashion-MNIST, CIFAR10 & No \\
  Detection with image residuals \cite{arvinte2020detecting} & VGG19 & Black-box attacks,$^p$ adaptive (white-box) attacks, gray-box attacks & CIFAR10 & No \\% $^q$
  % \hline
  % Inherent robustness 
  CompNet \cite{cosgrove2020robustness} & VGG16 \cite{simonyan2014very} & Black-box attacks$^p$ & Pascal3D+ \cite{xiang2014beyond}, GTSRB \cite{stallkamp2012man} & No \\
  Improved CapsNet \cite{deng2020detecting} & CNN, CapsNet \cite{hinton2018matrix} & White-box attacks$^p$ & MNIST, CIFAR10, SVHN & No \\
  PatchVeto \cite{huang2021zero} & ViT-B/16 \cite{wightman2019torch} & Adversarial patch & CIFAR10 \cite{krizhevsky2009learning}, ImageNet \cite{krizhevsky2012imagenet} & No \\
  \bottomrule
  \multicolumn{5}{L{.99\textwidth}}{\scriptsize$^o$The final pooling layer was excluded from ResNet9 used here, and the presented certified accuracy included that with and without affine transformation of the adversarial patch.}\\
  \multicolumn{5}{L{.99\textwidth}}{\scriptsize$^p$To be specific, PGD (Projected Gradient Descent) attack \cite{madry2017towards}, C\&W attack \cite{carlini2017towards}, brute-force PGD attack, and single-pixel attack \cite{su2019one} were used in \cite{arvinte2020detecting}; Texture PatchAttack \cite{yang2020patchattack} and the modified SparseRS \cite{croce2020sparse} were used in \cite{cosgrove2020robustness}; Naive white-box attack and adaptive white-box attack were used in \cite{deng2020detecting}.}\\
\end{tabular}
}}%
\end{table*}

% \subsection{Defenses based on small receptive fields}
\subsection{Defences based on small receptive fields}
\label{subsec:defense,receptive}

Xiang \etal{} \cite{xiang2021patchguard} presented a defensive technique against adversarial patches named \textit{PG (PatchGuard)} that used CNNs with small receptive fields to build robust classifiers. 
A small receptive field restricted the number of features that were influenced by the attack and helped to find feature boundaries. 
After that, a feature aggregation method was used to mask and recover the correct prediction. 
PG was claimed not only to be robust but also to maintain high clean accuracy against localised adversarial patches. 
Later on, Xiang and Mittal \cite{xiang2021patchguard++} proposed \textit{PG++ (PatchGuard++)} to detect patch attacks by feature extraction. 
They demonstrated that PG++ significantly improved not only the provable robust accuracy but also the clean accuracy. 
Furthermore, Xiang and Mittal \cite{xiang2021detectorguard} also presented \textit{DG (DetectorGuard)}, which first achieved a provable robustness against hidden localised patches, providing formal guarantees in an adversarial setting. 
% the first provable robustness against hidden localised patches by providing formal guarantees in an adversarial setting. 
It aimed to secure object detectors mainly in autonomous driving, video surveillance, and identity verification, while most of the adversarial defences focused on the domain of image classification.

% \subsection{Certified defenses}
\subsection{Certified defences}
\label{subsec:cd}

Most defences proposed in the context of adversarial patches are based on pre-processing of inputs at inference time \cite{hayes2018visible,naseer2019local}. 
However, Chiang \etal{} \cite{chiang2020certified} argued that these defences were easy to break with white-box adversaries, which encouraged the development of certified defences. 
Certified defences could not only defend against patch attacks but also provided the guaranteed confidence with what they were able to defend. 
Certified defences are usually computational expensive as they involve evaluation of extreme bounds during operations that help to certify the model given the worst-case scenarios. 
Therefore, they are relatively slower and less scalable than their counterparts. 
Despite the limitations of time cost, certified defences are the first step towards the ultimate robustness in deep learning based vision systems. 
We discuss the defence techniques that ensure certification in this subsection.

% ExtIBP \textit{(an extension of IBP to patches)}
Chiang \etal{} \cite{chiang2020certified} proposed the first certified defence against patch attacks with CertIBP \textit{(IBP certified models)} and faster ways of training, whereas the former shared some similarities with \textit{IBP (Interval Bound Propagation)} \cite{gowal2018effectiveness} and \textit{CROWN-IBP} \cite{zhang2019towards}. 
They observed the certified accuracy of CertIBP outperformed the empirical accuracy of LGS \cite{naseer2019local} and DW \cite{hayes2018visible} and that the certified defence was robust against all shapes of an adversarial patch. 
Despite the substantial guarantee over the certified defence, however, its scalability was demanding and infeasible due to the computational burden increased quadratically with the size of images. 
%% because the computational ..
Levine and Feizi \cite{levine2020randomized} utilised randomised smoothing---which was extensively used for certified defences against $L_2$ \cite{cohen2019certified}, $L_1$ \cite{lecuyer2019certified}, and $L_0$ \cite{levine2020robustness} attacks---and extended robustness through it to \textit{DRS (De-Randomised Smoothing)} for certified defences against patch attacks. 
Specifically, they exploited the fact that patch attacks appear in a more constrained setting than $L_0$ attacks \cite{levine2020robustness} and accordingly adopted a de-randomised procedure to craft defences that incorporated knowledge of the patch structure. 
Although this was a pure defence strategy that did not reveal the patch location, it did provide guaranteed robustness against generic patch attacks. 
%%
% a certifiable defense with \textit{randomised cropping}
% provable patch defense with randomised cropping
% RCD, RandCrop, RC
Moreover, Lin \etal{} \cite{lin2021certified} proposed RCD \emph{(Randomised Cropping Defence)} against patch attacks, where random cropped subsets from the original image were classified independently and the original image was classified as the majority vote over predicted classes of the sub-images. 
They also claimed certified robustness bounds for the model, leveraging the fact that a patch attack can only affect a certain number of pixels in the image. 
Their proposed method showed comparable clean accuracy, faster inference time, and higher certified accuracy under the worst-case scenarios than DRS \cite{levine2020randomized} and PG (PatchGuard) \cite{xiang2020patchguard}.

% \textit{BAG-CERT} % The model in BAG-CART
Ideally, certified defences should have certification as part of their training objectives, as certification done after training often requires post-hoc calibration. 
Metzen and Yatsura \cite{metzen2021efficient} proposed \emph{BagCert} inspired by BagNet \cite{brendel2019approximating}, combining a specific model architecture with a certified training procedure, to scale to larger patches and provide higher accuracy than CertIBP \cite{chiang2020certified}. 
The BagCert model could achieve good accuracy even with small receptive fields which helped in reducing the regions affected by the adversarial patch in the final feature map. 
BagCert outperformed CertIBP in terms of certified accuracy with more scalability, and its certification time was relatively low on CIFAR10 (43 seconds for 10,000 images) and ImageNet (7 minutes for 50,000 images) with higher certified accuracy than the other certified defences.
There are also some additional certified defences (like DRS \cite{levine2020randomized}, PG \cite{xiang2020patchguard}, and CROWN-IBP \cite{zhang2019towards}) which involved inference time computations instead of actual model training for certified robustness.

\iffalse%

% \subsection{Defenses based on small receptive fields}
\subsection{Defences based on small receptive fields}
\label{subsec:defense,receptive}

Xiang \etal{} \cite{xiang2021patchguard} presented a defensive technique against adversarial patches named \textit{PG (PatchGuard)} that used CNNs with small receptive fields to build robust classifiers. 
A small receptive field restricted the number of features that were influenced by the attack and helped to find feature boundaries. 
After that, a feature aggregation method was used to mask and recover the correct prediction. 
PG was claimed not only to be robust but also to maintain high clean accuracy against localised adversarial patches. 
%%
Later on, Xiang and Mittal \cite{xiang2021patchguard++} proposed \textit{PG++ (PatchGuard++)} to detect patch attacks by feature extraction. 
They demonstrated that PG++ significantly improved not only the provable robust accuracy but also the clean accuracy. 
%%
Furthermore, Xiang and Mittal \cite{xiang2021detectorguard} also presented \textit{DG (DetectorGuard)}, the first provable robustness against hidden localised patches by providing formal guarantees in an adversarial setting. 
It aimed to secure object detectors mainly in autonomous driving, video surveillance, and identity verification, while most of the adversarial defences focused on the domain of image classification. 

\fi%

% \subsection{Domain specific defenses}
% \label{subsec:asd}

% \subsection{Other defense techniques}
\subsection{Other defence techniques}
\label{subsec:od}

Most defence techniques adopted inference time methodology/implementation because adversarial training for patch attacks is an inference task. 
McCoyd \etal{} \cite{mccoyd2020minority} proposed \textit{MR (Minority Reports defence)} that used occlusions to adversarially train the model and then detect and defend against the adversarial patch at inference time. 
% to incorporate them both by using occlusions, adversarially training the model and then at inference time detecting and defending against the adversarial patch. 
%%
\iffalse%
In the defense design, they initially applied occlusion patches at different locations in each image with a fixed stride and trained the model, ensuring that the model could correctly predict occluded images. 
After that, the occlusion was carried out over the attacked image when images were attached with an adversarial patch,  traversing each pixel with the fixed stride and computing its confidence scores of prediction to form a prediction grid. 
They also provided a worst case security analysis against adaptive attacks. 
\fi%
Although MR was shown to have better clean and certified accuracy\footnote{%
compared with CertIBP \cite{chiang2020certified}
} and more effectiveness against adaptive attacks,\footnote{%
unlike DW \cite{hayes2018visible} and LGS \cite{naseer2019local}
} McCoyd \etal{} \cite{mccoyd2020minority} made a few assumptions that might be infeasible in practical scenarios. 
For example, in order to train the model using occlusions with an appropriate size, they assumed that the size of patch attacks was known. % given. 
Even if a soft agreement policy was adopted instead of hard unanimity to report defences in the prediction grid, the defence robustness still depended on the accuracy of the model on occluded images.

Arvinte \etal{} \cite{arvinte2020detecting} proposed a \textit{two-stage detection process using image residuals} for patch-based attacks. 
They claimed that it had better generalisation\footnote{%
compared to a \textit{detection baseline using prediction probabilities} \cite{hendrycks2016baseline} and \textit{LID (Local Intrinsic Dimensionality)} \cite{ma2018characterizing}
} and effectiveness for the detection against patch adversarial attacks.\footnote{%
like \textit{PGD (Projected Gradient Descent)} \cite{madry2017towards}, \textit{norm-restricted C\&W} ($L_0$, $L_2$, and $L_\infty$) \cite{carlini2017towards}, and \textit{one-pixel} \cite{su2019one} attacks} 
This generalised detection technique could detect strong black-box attacks, resist transfer attacks, and decrease the success rate of white-box attacks. 
However, they didn't address the introduced latency due to an added detection classifier. 
Co \etal{} \cite{co2021real} proposed \textit{HyperNeuron} to allow real-time detection for \textit{UAPs (Universal Adversarial Perturbations)} \cite{moosavi2017universal} by identifying suspicious neuron hyper-activations. 
It could provide the defence against adversarial masks and patch attacks simultaneously with significantly lower latency, making it usable in real-time applications. 
However, limitations existed because they only chose the mean and standard deviation as aggregation functions, ignoring other statistical measures that should be considered as part of the evaluation.

\subsection{Domain specific defences}
\label{subsec:asd}
% defenses

Since adversarial patch attacks can be practically applied in diverse scenarios, some domain-specific defences are also proposed to cater to the domain-specific application requirements. 
For example, Pavlitskaya \etal{} \cite{pavlitskaya2020feasibility} extended LGS \cite{naseer2019local} and investigated the affected patch performance and effectiveness under various conditions (such as boundaries, weather, and lighting) in an end-to-end vehicle control situation. 
They found that salient regions were edges of the road and lane markings while adversarial patches depended on the environment and closed/open loop settings, highlighting the constraints and boundary conditions worth exploiting to suppress noise. 
Furthermore, they observed that LGS-based defences helped to build a robust model against attacks via saliency maps generated by VisualBackProp \cite{bojarski2017explaining,bojarski2016visualbackprop}.

Nesti \etal{} \cite{nesti2022evaluating} evaluated adversarial patch attacks in semantic segmentation based applications like autonomous driving, while most adversarial patch attacks focus on either classification or detection. 
They presented a \textit{scene-specific attack} for autonomous driving scenarios and demonstrated the robustness of state-of-the-art models against patch attacks. 
They found that the patch attack trained by the \textit{EOT (Expectation Over Transformation)} \cite{athalye2018synthesizing} strategy proved more robust than the normal adversarial patches in both 3D CARLA virtual simulations and real-world experiments, but the latter performed better on the Cityscapes \cite{cordts2016cityscapes} dataset. 
Although the addition of patches did reduce the model accuracy, Nesti \etal{} \cite{nesti2022evaluating} observed a contrasting behaviour compared to classification and detection tasks, as the semantic model was not easily corrupted by adversarial patches.

Moreover, L\o{}kken \etal{} \cite{lokken2020investigating} discussed the effectiveness of GAN-generated adversarial camouflages for a wide range of DNN classifiers of naval vessels. 
They trained a GAN to generate adversarial masks that were placed on the images later and found that the DNN classifier was indeed possibly weakened. 
However, they trained and tested the adversarial camouflage only on grayscale images, while most images collected nowadays through remote sensing and drone surveillance are in color. 
Therefore, the robustness of this technique still needs to be evaluated for colored images.

\subsection{Inherent robustness}
\label{subsec:defense,robust}

Most of the empirical defences for adversarial patches are based on either adversarial training or on saliency map inference. 
A different approach was discussed by Cosgrove \etal{} \cite{cosgrove2020robustness}. % The authors 
They utilised the fact that adversarial patches are maximally difficult occlusions and accordingly proposed \emph{CompNets}, an interpretable compositional model, to effectively defend against adversarial patches. 
Their models were inherently robust to occlusions, avoiding the need for adversarial training, and have been extensively used to produce robust models against normal occlusions in scenes \cite{kortylewski2020compositional,kortylewski2021compositional,kortylewski2020combining}. 
Although the interpretability of CompNets made them attractive as defences, the robustness of these models could be hindered by fine-tuning or combining conventional CNN models, while their performance was similar to their counterparts based on adversarial training.

% \textit{CapsNets (capsule networks)}
% the effectiveness of CapNets
% They also evaluated its effectiveness in classifying and detecting adversarial patched inputs and proposed two modifications to the CapsNet architecture (\ie{} \textit{affine voting} and \textit{matrix capsule dropout}) to enhance classification accuracy. 
Subsequently, Deng \etal{} \cite{deng2020detecting} proposed an \emph{Improved CapsNet}, which was one of the earliest detection methods after SentiNet \cite{chou2020sentinet}, exploring the effectiveness of adversarial patch detection methods. 
They proposed two modifications to the CapsNet (Capsule Network) \cite{hinton2018matrix} architecture (\ie{} \textit{affine voting} and \textit{matrix capsule dropout}) to enhance classification accuracy and also evaluated its effectiveness in classifying and detecting adversarial patched inputs. 
However, the attack methods used to highlight the effectiveness of evaluation were naive whitebox attacks and adaptive whitebox attacks, which are limitations of this technique. 
Blackbox attack methods should have also been considered as they are the most practical use case scenarios for patch attacks. 
Moreover, latency was introduced during classification due to the presence of detection methods yet not discussed. 
If latency was high, the detection method would ultimately prove not to be helpful for real-world scenarios.

Huang and Li \cite{huang2021zero} proposed \textit{PatchVeto}, a zero-shot certified defence against adversarial patches based on ViT (Visual Transformers) \cite{dosovitskiy2020image} models, requiring no prior training. 
PatchVeto closely resembled MR \cite{mccoyd2020minority}, but MR cannot scale to higher resolution datasets like ImageNet because of the training difficulty and heavy computation overhead. 
% the latter did not
PatchVeto showed significant performance improvement on the high-resolution ImageNet \cite{deng2009imagenet} dataset, while MR \cite{mccoyd2020minority} and CertIBP \cite{chiang2020certified} were either difficult or infeasible to implement. 
PatchVeto also showed higher clean accuracy than the BagNet-based defences and higher certified accuracy than other certified defences \cite{xiang2020patchguard,mccoyd2020minority,metzen2021efficient,levine2020randomized,chiang2020certified}, because it could capture global features in the scene as well as relations among different regions thanks to its attention mechanism. 
However, PatchVeto was only demonstrated against square patches while patches of other shapes could affect its declared performance. 
Moreover, the system latency might be slightly higher as each image was verified against a batch of images whose regions were masked during inference. 
But considering the design of the zero-shot defence, the latency could be improved through various extensions in the future. 
%% beauty

Salman \etal{} \cite{salman2021certified} presented certified patch attacks by leveraging ViTs to achieve significantly higher robustness guarantees and maintain standard accuracy and inference time capabilities comparable to non-robust models. 
However, they only used column ablations to certify patch defences. 
Lennon \etal{} \cite{lennon2021patch} proposed \textit{mAST (mean Attack Success over Transformations)} based on mean average robustness as a new metric to evaluate the robustness and invariance of patch attacks. 
The robustness of patch attacks was evaluated for 3D positions and orientations under various conditions, and it was qualitatively demonstrated that for some 3D transformations, increasing the training distribution support could increase the patch success. 
However, mAST was only evaluated on whitebox models, while blackbox models should have been taken into account for evaluation purposes as they are more practically attainable. 
% realisable

% \section{Discussion and conclusion}
\section{Conclusion}
\label{sec: conclusion}

Adversarial patch attacks are an open and active field of research, with plentiful attack and defence methods continuously emerging every year. 
Here, we presented a clear and comprehensive survey of existing techniques for adversarial patch attacks and defences focusing on vision-based tasks, intending to give a strong foundation for those wanting to understand their capabilities and weaknesses. 
However, most adversarial publications still remain in the classification and object detection fields, while there are plenty of challenges and other application scenarios that exist in the domain, such as scalability, real-time capability, and many others. 
Other applications of adversarial patch attacks in language or translation models might be an interesting direction as well. 
Moreover, given the non-explainability of the DNN-based blackbox models, would adversarial patch attacks be able to provide a new perspective in  interpretation of model predictions? If so, could they be used to help build more powerful models in real-world scenarios? 
All of these questions are really exciting and worth the effort to investigate, and we genuinely look forward to seeing future solutions get promoted to make progress in this field and further benefit our society.

\bibliographystyle{IEEEtran}% plain}
% \bibliography{ref} % iso4=abbr
% \bibliography{second_full,second_refs}
\bibliography{second_abbr,second_refs}

%

\end{document}